\documentclass{article}

\usepackage[preprint]{neurips_2026}

\usepackage[utf8]{inputenc} 
\usepackage[T1]{fontenc}    
\usepackage{hyperref}       
\usepackage{url}            
\usepackage{booktabs}       
\usepackage{amsfonts}       
\usepackage{amsmath}        
\usepackage{nicefrac}       
\usepackage{microtype}      
\usepackage{xcolor}         
\usepackage{tikz}           
\usetikzlibrary{shapes.geometric, arrows.meta, positioning, calc, fit, backgrounds, decorations.pathreplacing}
\usepackage{algorithm}      
\usepackage{algorithmic}    
\usepackage{multirow}       
\usepackage{adjustbox}      

\title{When Agents Evolve, Institutions Follow}

\author{%
  Chao Fei \\
  KAUST\\
  \texttt{chao.fei@kaust.edu.sa} \\
  \And
  Hongcheng Guo \\
  Fudan University \\
  \And
  Yanghua Xiao \\
  Fudan University \\
}

\begin{document}

\maketitle

\begin{abstract}

Across millennia, complex societies have faced the same coordination problem of how to organize collective action among cognitively bounded and informationally incomplete individuals. Different civilizations developed different political institutions to answer the same basic questions of who proposes, who reviews, who executes, and how errors are corrected. We argue that multi-agent systems built on large language models face the same challenge. Their central problem is not only individual intelligence, but collective organization. Historical institutions therefore provide a structured design space for multi-agent architectures, making key trade-offs between efficiency and error correction, centralization and distribution, and specialization and redundancy empirically testable. We translate seven historical political institutions, spanning four canonical governance patterns, into executable multi-agent architectures and evaluate them under identical conditions across three large language models and two benchmarks. We find that governance topology strongly shapes collective performance. Within a single model, the gap between the best and worst institution exceeds 57 percentage points, while the optimal architecture shifts systematically with model capability and task characteristics. These results suggest that collective intelligence will not advance through a single optimal organizational form, but through governance mechanisms that can be reselected and reconfigured as tasks and capabilities evolve. More broadly, this points to a transition from \textbf{self-evolving agents} to the \textbf{self-evolving multi-agent system}. The code is available on \href{https://github.com/cf3i/SocialSystemArena}{GitHub}.
\end{abstract}

\section{Introduction}

Since antiquity, every society of sufficient scale has faced the same coordination problem. It must organize collective action among individuals who are cognitively bounded and informationally incomplete, and there is no single best solution to this problem \citep{olson1965}. Every such society must decide who may propose solutions, who reviews them, who executes them, and how errors are detected and corrected. Across civilizations and historical periods, these questions have received different answers. Those answers gave rise to coherent governance architectures, including centralized hierarchies, layered review systems, autonomous federations, and consensus-based democracies. Each represents a distinct response to collective action under bounded rationality.

This diversity is not accidental. It reflects structured choices among competing solutions to the same underlying coordination problem under different constraints, risks, and capabilities. History offers no evidence that one political form is universally superior. Different institutions make different trade-offs among efficiency, robustness, error tolerance, and adaptability, and those trade-offs shape whether they persist or disappear over time. We therefore treat historical institutions not as literal equivalents of multi-agent systems, but as historically tested governance templates that instantiate recurring solutions to coordination under bounded rationality. This makes them a structured design space for studying how governance topology shapes collective performance.

A long line of organizational research points to a simple conclusion. In systems composed of bounded-rational agents engaged in collective action, structure matters in its own right, and governance topology strongly shapes outcomes \citep{marchsimon1958, woodward1965, thompson1967, ostrom1990, north1990, granovetter1985, powell1990}. Multi-agent systems built on large language models fit this description closely. Their agents operate with limited context, partial information about global state, and finite reasoning capacity, so they must coordinate through some governance arrangement \citep{guo2024llm_multiagent_survey}. In that sense, they face a coordination problem that is structurally similar to the one long faced by human institutions.

This matters for two reasons. First, single-agent systems face a real ceiling on complex tasks. Limited context, finite deliberation depth, and incomplete access to distributed information make one agent brittle when it must plan, review, and act at the same time. Second, as self-evolving agents become more capable and more autonomous~\citep{liang2026genericagent, nousresearch2025hermesagent}, monolithic designs become harder to trust. Concentrating information, decision authority, and execution power in one agent raises sharper problems of privacy exposure, unsafe action, weak oversight, and poor failure recovery. Multi-agent governance is therefore needed not only to extend capability beyond the limit of a single agent, but also to separate roles, constrain authority, and support structured review and correction.

\begin{figure}[t]
\centering
\includegraphics[width=0.75\textwidth]{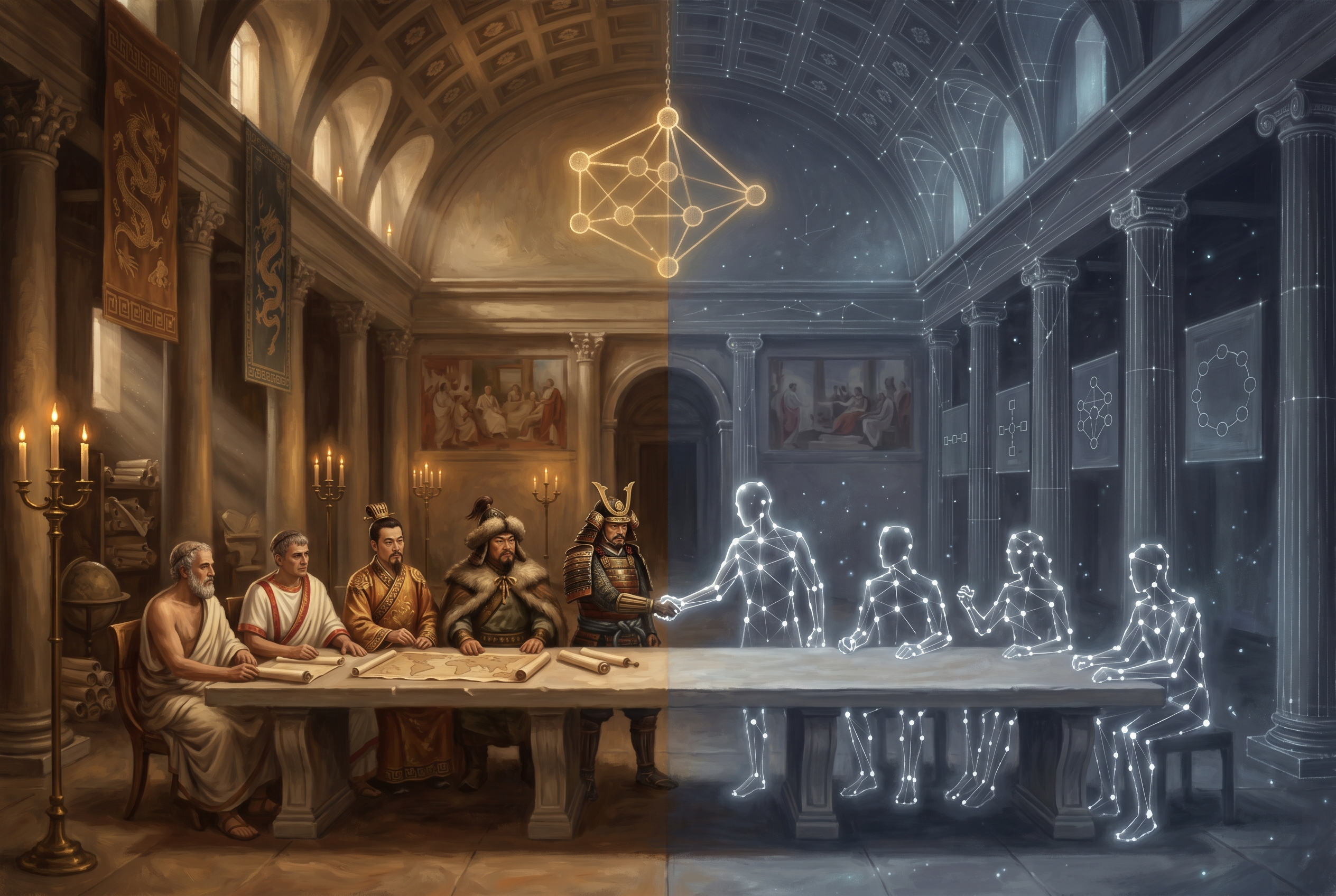}
\caption{Historical institutions as a design space for multi-agent governance. Millennia of political experimentation provide a structured space of governance topologies, while controlled study of these topologies in multi-agent systems can help reveal their trade-offs more clearly.}
\label{fig:homology}
\end{figure}

Despite rapid progress in multi-agent systems, most existing work focuses on role prompting, debate, planning, or tool use, while giving much less attention to governance topology as the central design variable. We still lack a controlled framework for comparing different governance structures under matched conditions and for asking when one topology works better than another. Historical institutions offer a natural starting point for this question. They are governance solutions that have already been selected, adapted, and stress-tested in real collective systems. They can now be studied as alternative architectures under controlled experimental conditions (Figure~\ref{fig:homology}).

In this paper, we recast seven historical political institutions as executable multi-agent architectures and evaluate them under unified experimental conditions. These institutions span the four canonical governance modes introduced above. We compare them across three large language models and two benchmarks with different task characteristics, which allows us to identify how governance topology interacts with model capability and task structure. In doing so, we turn governance from an intuitive analogy into a reproducible empirical object.

Our results show that governance topology is a major determinant of multi-agent system performance. On the same model, the gap between the best and worst institution exceeds 57 percentage points. Yet the best topology is not fixed. It shifts systematically with model capability and task characteristics. No single institution dominates across all models and tasks, which echoes the historical record in which no universal political form prevailed across societies and environments. Architectural choice should therefore be treated as a system design variable on par with model capability. More broadly, collective intelligence is unlikely to improve through a single best organization. What matters is whether governance can be re-selected and reconfigured as tasks, capabilities, and risk conditions change. In the longer run, this motivates a shift from studying isolated self-evolving agents to studying agent societies that can adapt their governance over time.

\begin{enumerate}
    \item \textbf{Executable governance framework.} We introduce \textsc{SocialSystemArena}, a formal framework that translates historically grounded political institutions into computable governance specifications $\mathcal{G} = (P, A, S, T, F)$. This makes governance topology an explicit and isolatable design variable for multi-agent systems.

    \item \textbf{Controlled topology benchmark.} We build a unified \textsc{GovernanceRuntime} and use it to compare seven governance architectures under matched conditions across three large language models and two benchmarks. This makes it possible to measure how institutional structure shapes collaboration quality, efficiency, robustness, and failure patterns.

    \item \textbf{Adaptive governance for self-evolving agent systems.} We show that no single governance form dominates across models and tasks. Instead, governance optima shift systematically with model capability and task structure. We further identify gate density $\rho$ as a useful predictor of governance overhead, characterize the gate-loop failure mode, and provide empirical support for adaptive topology selection as a core design principle for self-evolving agent systems.
\end{enumerate}

\section{Related Work}
\label{sec:related}

The performance of multi-agent systems is shaped not only by the capabilities of individual agents but also by how they are organized---their communication topology, decision flow, and coordination mechanisms. Yet this governance dimension remains poorly understood: how do different organizational structures affect MAS performance, and does the answer change with the underlying model and task? We review four lines of research that expose this gap.

\subsection{Static Multi-Agent Frameworks}
\label{sec:rw-static}

Recent LLM-based multi-agent frameworks have demonstrated that orchestrating specialized agents can outperform single-model baselines, each adopting a distinct communication pattern: multi-turn dialogue~\citep{wu2023autogen}, sequential pipelines~\citep{hong2023metagpt, qian2024chatdev}, role-playing pairs~\citep{li2023camel}, dynamic team recruitment~\citep{chen2024agentverse}, and adversarial debate~\citep{du2024debate}.
Although some frameworks support flexible agent wiring, each demonstrates a particular topology without systematically comparing alternatives under controlled conditions---leaving open whether the observed gains stem from the topology itself or from other design choices.

\subsection{Topology Optimization}
\label{sec:rw-topology}

A growing body of work explores automatic topology search. GPTSwarm~\citep{zhuge2024gptswarm} jointly optimizes node prompts and edge connectivity via gradient methods; G-Designer~\citep{zhang2025gdesigner} generates task-adaptive topologies with a variational graph auto-encoder; MacNet~\citep{qian2025macnet} identifies collaborative scaling laws using DAGs; EvoMAC~\citep{hu2025evomac} enables agents and connections to self-evolve at test time; and EvoAgent~\citep{yuan2024evoagent} applies evolutionary operators to grow diverse agent populations from a single seed.
These methods typically optimize edge-wise connectivity within homogeneous agent pools and validate on a single model backend, leaving open whether the discovered topologies generalize across LLMs and task distributions.

\subsection{LLM-based Social Simulation}
\label{sec:rw-social}

A complementary thread treats LLM agents as members of artificial societies. \citet{park2023generative} demonstrated that LLM agents exhibit rich social dynamics through autonomous planning and coordination, while GovSim~\citep{piatti2024govsim} showed that explicit institutional mechanisms are necessary for agents to achieve cooperation in commons-dilemma settings.
These studies observe what behavior emerges from a given agent population, but do not systematically vary the institutional structure to isolate its causal effect on task outcomes.

\subsection{Agent Evaluation Benchmarks}
\label{sec:rw-bench}

Systematic agent evaluation has advanced through AgentBench~\citep{liu2024agentbench} (eight diverse environments), AgentBoard~\citep{ma2024agentboard} (fine-grained progress metrics), ReAct~\citep{yao2023react} (interleaved reasoning-acting), Toolformer~\citep{schick2023toolformer} (tool acquisition), and rationality-oriented evaluations~\citep{jiang2025rationality}.
These benchmarks universally fix the governance structure and vary the underlying model, leaving governance topology as an uncontrolled---and therefore unstudied---variable.

\section{Methodology}
\label{sec:methodology}

\begin{figure}[t]
\centering
\includegraphics[width=\textwidth]{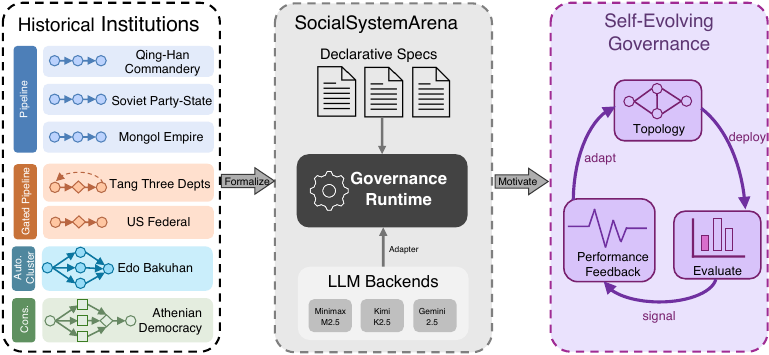}
\caption{Overview of \textsc{SocialSystemArena}. \emph{Left}: Seven historical institutions span four canonical governance patterns. \emph{Center}: Each institution is formalized as a declarative governance specification and executed on a unified runtime across three LLM backends and two benchmarks. \emph{Right}: A long-horizon vision of self-evolving multi-agent systems that reconfigure governance topology as tasks and model capabilities evolve.}
\label{fig:overview}
\end{figure}

We present \textsc{SocialSystemArena}, a framework that derives multi-agent governance architectures from historical political institutions and evaluates them on a unified runtime (Figure~\ref{fig:overview}).

\subsection{Problem Formulation}
\label{sec:formulation}

We formalize a multi-agent governance architecture as a \emph{Governance Specification} $\mathcal{G} = (P, A, S, T, F)$, where:
\begin{itemize}
    \item $P \in \{\texttt{pipeline},\; \texttt{gated\_pipeline},\; \texttt{autonomous\_cluster},\; \texttt{consensus}\}$ is the \emph{pattern}, defining the message-flow topology class.
    \item $A = \{a_1, \dots, a_n\}$ is a set of agents. Each agent $a_i$ is assigned a role $r_i$ drawn from a fixed role vocabulary (e.g., planner, gatekeeper, executor) and a \emph{soul prompt} $\sigma_i$ that encodes its institutional persona and behavioral constraints.
    \item $S = (s_1, \dots, s_m)$ is an ordered sequence of \emph{stages}. Each stage $s_j$ is bound to an agent (or a set of voters/cluster members) and specifies a \texttt{kind} aligned with its agent's role.
    \item $T: S \times D \to S$ is the \emph{transition function}, where $D$ is the decision space of string-valued routing decisions output by agents (e.g., \texttt{next}, \texttt{approve}, \texttt{reject}).
    \item $F = \{f_1, \dots, f_k\}$ is a set of \emph{features}, pluggable behavioral modifiers orthogonal to the pattern (e.g., monitoring, shared state propagation, policy enforcement).
\end{itemize}

Given a task $\tau$, the runtime executes $\tau$ under governance $\mathcal{G}$ by stepping through stages according to $T$, producing a trace of decisions, summaries, and artifacts.

Our investigation decomposes into two questions. First, holding the LLM backend and task fixed, how does the governance specification $\mathcal{G}$ affect task completion rate, token consumption, and execution step count? Second, does the optimal $\mathcal{G}$ shift systematically across LLM backends and task structures? The first question isolates governance topology as an architectural variable; the second tests whether that variable interacts with model capability and task characteristics.

\subsection{Four Canonical Patterns}
\label{sec:patterns}

We distill four canonical message-flow patterns from historical governance systems, each abstracting a distinct organizational logic. Each pattern imposes compile-time constraints on the allowed stage kinds, enforced by the specification validator before runtime.

\begin{figure}[t]
\centering
\includegraphics[width=0.85\textwidth]{assets/patterns.png}
\caption{Four canonical governance patterns. (a)~\textbf{Pipeline}: linear single-direction flow. (b)~\textbf{Gated Pipeline}: pipeline with gate stages that can reject and loop back. (c)~\textbf{Autonomous Cluster}: orchestrator dispatches to parallel subsystems. (d)~\textbf{Consensus}: proposer triggers parallel voting; tally determines flow.
Circles denote stages, diamonds denote gate stages, double circles denote terminal, cyan boxes denote cluster members, green boxes denote voters.}
\label{fig:patterns}
\end{figure}

\paragraph{Pipeline.}
Stages form a linear chain $s_1 \to s_2 \to \cdots \to s_m$; each stage produces a single \texttt{next} decision advancing to the successor. No branching, gating, or parallel execution occurs.

\paragraph{Gated Pipeline.}
Extends the pipeline with \emph{gate stages} whose transitions include both an \texttt{approve} path (advancing forward) and a \texttt{reject} path (looping back to a prior stage for revision), forming an error-correction cycle. We define the \emph{gate density} $\rho = g / |S|$, where $g$ is the number of gate stages and $|S|$ the total stage count including terminal. $\rho$ captures how much of the topology is devoted to review rather than execution, and we analyze it as a predictor of governance overhead in \S\ref{sec:eval}. A \emph{gate-loop failure} occurs when repeated \texttt{reject} cycles exhaust the step budget $B$ without reaching the terminal stage, a failure mode we characterize empirically in \S\ref{sec:eval}.

\paragraph{Autonomous Cluster.}
An orchestrator distributes work to autonomous subsystems that execute \emph{in parallel}; an aggregation stage collects results and emits \texttt{success} if all required members succeed or \texttt{failure} otherwise.

\paragraph{Consensus.}
A proposer formulates a proposal submitted to voters dispatched \emph{in parallel}. Votes are aggregated against a threshold $\theta$ using a configurable rule (majority, weighted, or unanimity).

Figure~\ref{fig:patterns} illustrates the four patterns; Table~\ref{tab:institutions} lists which institutions instantiate each pattern and their structural parameters.

\subsection{Governance Runtime}
\label{sec:runtime}

All institutions share a single \textsc{GovernanceRuntime} implementation. The runtime code, LLM adapter, and model backend are identical for every institution, guaranteeing that the governance specification $\mathcal{G}$ is the \emph{only controlled variable}. Algorithm~\ref{alg:runtime} presents the core execution loop. For each step, the runtime resolves the current stage, dispatches it by kind (single agent, consensus, or cluster), applies feature plugins, and follows the transition function $T$ to advance.

\begin{algorithm}[t]
\caption{\textsc{GovernanceRuntime.Run}($\mathcal{G}$, $\tau$, $B$)}
\label{alg:runtime}
\begin{algorithmic}[1]
\REQUIRE Governance spec $\mathcal{G}$; task $\tau$; step budget $B$
\ENSURE \textsc{TaskState} with execution trace
\STATE $\textit{task} \leftarrow \textsc{InitTask}(\tau,\; \mathcal{G}.\text{entry\_stage})$
\FOR{$\textit{step} = 1$ \TO $B$}
    \STATE $\textit{stage} \leftarrow \mathcal{G}.\text{stage\_map}[\textit{task}.\text{current\_stage}]$
    \IF{$\textit{stage}.\text{kind} = \texttt{terminal}$}
        \STATE $\textit{task}.\text{status} \leftarrow \texttt{done}$; \textbf{break}
    \ENDIF
    \STATE $\textit{ctx} \leftarrow \textsc{BuildContext}(\textit{task}, \textit{stage})$
    \FORALL{$f \in F$} \STATE $f.\textsc{BeforeStage}(\textit{task}, \textit{stage}, \textit{ctx})$ \ENDFOR
    \IF{$\textit{stage}.\text{kind} = \texttt{consensus}$}
        \STATE $\textit{result} \leftarrow \textsc{RunConsensus}(\textit{stage}, \textit{ctx})$ \COMMENT{parallel voter dispatch + aggregation}
    \ELSIF{$\textit{stage}.\text{kind} = \texttt{cluster}$}
        \STATE $\textit{result} \leftarrow \textsc{RunCluster}(\textit{stage}, \textit{ctx})$ \COMMENT{parallel member dispatch}
    \ELSE
        \STATE $\textit{result} \leftarrow \textsc{DispatchSingle}(\textit{stage}, \textit{ctx})$
    \ENDIF
    \FORALL{$f \in F$} \STATE $\textit{result} \leftarrow f.\textsc{AfterStage}(\textit{task}, \textit{stage}, \textit{result}, \textit{ctx})$ \ENDFOR
    \STATE $d \leftarrow \textit{result}.\text{decision}$ \COMMENT{overrides and defaults handled in App.~\ref{app:runtime}}
    \STATE $\textit{next} \leftarrow T(\textit{stage},\, d)$
    \STATE $\textit{task}.\text{history}.\text{append}\bigl(\textsc{Event}(\textit{stage}, d, \textit{next})\bigr)$
    \STATE $\textit{task}.\text{current\_stage} \leftarrow \textit{next}$
\ENDFOR
\RETURN $\textit{task}$
\end{algorithmic}
\end{algorithm}

The runtime further comprises a layered prompt assembly mechanism, six composable feature plugins (e.g., monitoring, loop guards, shared state propagation), and a unified LLM adapter protocol; details are provided in Appendix~\ref{app:runtime}.

\subsection{Institution Modeling}
\label{sec:institutions}

We translate seven historically representative institutions into executable governance specifications, spanning all four patterns plus a single-agent baseline (Table~\ref{tab:institutions}). Each translation follows four steps. In \emph{governance principle extraction} (step 0), we identify the core organizational wisdom that the historical institution embodies and formulate a testable hypothesis about its expected effect in a MAS context. In \emph{topology extraction} (step 1), we derive the information-flow structure from historical sources and map it to a canonical pattern. In \emph{role mapping} (step 2), we map historical roles to agent roles and stage kinds. In \emph{soul prompt authoring} (step 3), we encode institutional persona, behavioral constraints, and output format in a per-stage markdown file.

We summarize the governance principle and hypothesized MAS effect for each pattern family below.
\begin{itemize}
    \item \textbf{Pipeline institutions} (Qin-Han, Soviet, Mongol): \emph{Linear delegation through specialized stages.} Each stage handles a narrow role and passes results forward. We hypothesize that longer pipelines improve quality through division of labor, but with diminishing returns as chain length grows.
    \item \textbf{Gated-pipeline institutions} (Tang, US Federal): \emph{Centralized review as error correction.} Gate stages reject substandard outputs for revision, preventing error propagation. We hypothesize that this mechanism improves task completion when the LLM can satisfy the gate's criteria, but risks runaway revision loops when it cannot.
    \item \textbf{Autonomous cluster} (Edo): \emph{Decentralized parallel execution.} Independent subsystems operate autonomously and aggregate results. We hypothesize that this pattern maximizes throughput by eliminating sequential bottlenecks, at the cost of limited cross-agent coordination.
    \item \textbf{Consensus} (Athens): \emph{Democratic deliberation for collective decision-making.} Multiple voters evaluate proposals in parallel, reducing single-point-of-failure risk. We hypothesize that voting improves decision robustness but incurs coordination overhead that may outweigh its benefit in small-scale MAS.
\end{itemize}

\begin{table}[t]
\caption{Structural parameters of the eight governance specifications (7 institutions + 1 baseline). \emph{Gates}, \emph{Cluster} and \emph{Voters} count role-specific elements. $\rho$ is gate density. \emph{Monitor} indicates whether a side-channel observer agent is enabled.}
\label{tab:institutions}
\centering
\begin{adjustbox}{max width=\textwidth}
\begin{tabular}{@{}llcccccccc@{}}
\toprule
Institution & Pattern & Stages & Agents & Gates & Cluster & Voters & $\rho$ & Monitor \\
\midrule
Single Agent System (baseline) & pipeline & 2 & 1 & 0 & --- & --- & 0 & --- \\
Qin-Han Commandery-County & pipeline & 5 & 5 & 0 & --- & --- & 0 & \emph{yushi} \\
Soviet Party-State & pipeline & 6 & 5 & 0 & --- & --- & 0 & --- \\
Mongol Empire & pipeline & 7 & 6 & 0 & --- & --- & 0 & --- \\
\midrule
Tang Three Departments & gated pipeline & 6 & 10 & 1 & 6 & --- & 0.17 & --- \\
US Federal & gated pipeline & 9 & 8 & 5 & --- & --- & 0.56 & --- \\
\midrule
Edo Bakuhan & auton.\ cluster & 5 & 8 & 0 & 4 & --- & 0 & \emph{metsuke} \\
\midrule
Athenian Democracy & consensus & 5 & 10 & 0 & --- & 7 & 0 & --- \\
\bottomrule
\end{tabular}
\end{adjustbox}
\end{table}

The \textbf{Single Agent System (SAS)} baseline uses a single executor without governance scaffolding. The Athenian Democracy case study is deferred to Appendix~\ref{app:athens}.

\paragraph{Tang Three Departments (618--907 CE).} This institution instantiates the gated-pipeline pattern.

\begin{figure}[h]
\centering
\includegraphics[width=0.85\textwidth]{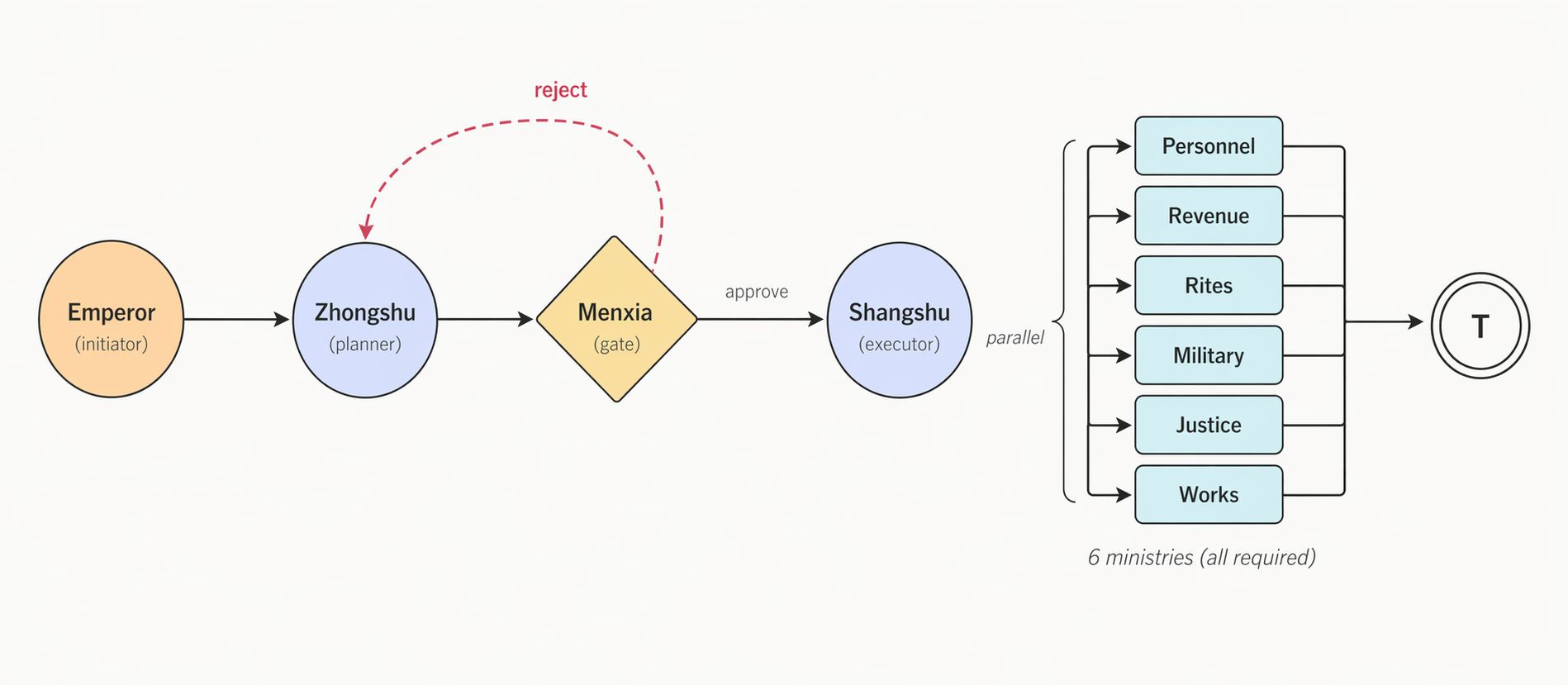}
\caption{Tang Three Departments topology. The Menxia gate's \texttt{reject} loops back to Zhongshu for revision, forming an error-correction cycle. The six ministries execute in parallel as a cluster stage.}
\label{fig:tang}
\end{figure}

The key structural feature is the \emph{Menxia gate}. Its \texttt{reject} decision loops back to the Zhongshu planner for revision, creating an error-correction cycle (an \texttt{imperial\_override} transition provides an escape path). The six ministries execute as a parallel cluster with all members required. This yields a low gate density ($\rho = 0.17$), devoting most of the topology to planning and execution.

The \textbf{Athenian Democracy} instantiates the consensus pattern with seven citizen-voters and majority aggregation; its topology and role details are provided in Appendix~\ref{app:athens}.

All institutions share one design principle. The governance specification $\mathcal{G}$ is the \emph{sole controlled variable}. Runtime code, LLM adapter, and model backend are identical across all experiments, and each institution's topology preserves historical fidelity.

\section{Evaluation}
\label{sec:eval}

We evaluate \textsc{SocialSystemArena} across two benchmarks, three LLM backends, and eight governance specifications. Our experiments address three questions: (1)~Does the choice of governance specification significantly affect task performance? (2)~Do institution rankings generalize across models and benchmarks? (3)~What is the cost--performance trade-off of structural complexity?

\subsection{Experimental Setup}
\label{sec:setup}

\paragraph{Benchmarks.}
We use two complementary evaluation suites.
\textbf{PinchBench} comprises 23 single-turn tool-use tasks spanning calendar management, web research, email drafting, file manipulation, data summarization, and multi-step API workflows. Each task is scored on a $[0,1]$ rubric with partial credit for sub-objectives.
\textbf{ClaweBench} comprises 104 multi-step real-world tasks organized into 24 categories across 6 difficulty levels, testing extended reasoning and multi-tool orchestration. Each task is scored on a $[0,1]$ rubric.

\paragraph{Models.}
We evaluate three commercially available LLM backends: \textbf{MiniMax M2.5}, \textbf{Kimi K2.5}, and \textbf{Gemini 2.5 Flash}. These were selected for API accessibility, cost feasibility, and provider diversity. All models are accessed through the unified adapter protocol (\S\ref{sec:runtime}); the governance specification is the only controlled variable.

\paragraph{Institutions.}
All eight governance specifications from Table~\ref{tab:institutions} are evaluated: seven historically grounded institutions plus the single-agent SAS baseline. Each (model, institution, benchmark) triple constitutes one full evaluation run.

\paragraph{Metrics.}
We report: \emph{Task Success Rate} (mean score across tasks), \emph{Average Steps} (mean runtime steps per task), \emph{Total Tokens} (aggregate token consumption), \emph{Zero-score Count} (number of tasks receiving a score of exactly 0), and \emph{Wall-clock Time}.

\subsection{Main Results on PinchBench}
\label{sec:pinch-results}

\subsubsection{Cross-Model Performance}

Table~\ref{tab:pinch-cross} reports the mean task success rate for each (model, institution) pair on PinchBench.

\begin{table}[t]
\caption{PinchBench task success rate (\%) across three LLM backends and eight governance specifications. \textbf{Bold}: best institution per model. \underline{Underline}: worst non-baseline institution per model.}
\label{tab:pinch-cross}
\centering
\begin{adjustbox}{max width=\textwidth}
\begin{tabular}{@{}llccc@{}}
\toprule
& Institution & MiniMax M2.5 & Kimi K2.5 & Gemini 2.5 Flash \\
\midrule
\multirow{3}{*}{\rotatebox[origin=c]{90}{\scriptsize Pipeline}}
& Single Agent System (baseline) & 31.1 & 21.7 & 63.8 \\
& Qin-Han Commandery-County & 79.5 & 30.2 & 65.3 \\
& Soviet Party-State & 61.6 & \underline{8.7} & 63.4 \\
& Mongol Empire & 59.3 & \textbf{67.3} & 46.3 \\
\midrule
\multirow{2}{*}{\rotatebox[origin=c]{90}{\scriptsize Gated}}
& Tang Three Departments & \textbf{88.2} & 21.6 & \underline{27.2} \\
& US Federal & \underline{43.0} & 13.0 & 51.1 \\
\midrule
& Edo Bakuhan (cluster) & 83.7 & 12.3 & \textbf{87.7} \\
\midrule
& Athenian Democracy (consensus) & 68.4 & 12.9 & 57.9 \\
\midrule
\multicolumn{2}{@{}l}{\emph{Cross-institution mean}} & 64.4 & 23.5 & 57.8 \\
\bottomrule
\end{tabular}
\end{adjustbox}
\end{table}

Three findings emerge.

\paragraph{Governance specifications substantially affect performance.}
Within a single model, the gap between the best and worst institution is large: 57.1 percentage points for MiniMax (Tang 88.2\% vs.\ Bare 31.1\%), 58.6 for Kimi (Mongol 67.3\% vs.\ Soviet 8.7\%), and 60.5 for Gemini (Edo 87.7\% vs.\ Tang 27.2\%). This confirms that the governance topology is a first-order determinant of multi-agent task performance, far exceeding the variance typically attributed to prompt engineering alone.

\paragraph{No universally optimal institution.}
The top-ranked institution differs across all three models: Tang for MiniMax, Mongol for Kimi, and Edo for Gemini. More strikingly, Tang---the overall best performer on MiniMax (88.2\%)---drops to near-worst on Gemini (27.2\%), while the simple SAS outperforms five multi-agent institutions on Gemini (63.8\%). This cross-model ranking instability suggests that institution--model compatibility is a critical factor that current governance design does not account for.

\paragraph{The gated-pipeline paradox.}
In principle, gate nodes should improve robustness by enabling error correction---rejected outputs are revised and resubmitted rather than propagated. In practice, both gated-pipeline institutions exhibit the \emph{highest} model sensitivity of any pattern. Tang ($\rho = 0.17$, single gate) achieves the best score on MiniMax but collapses on Kimi and Gemini, while US Federal ($\rho = 0.56$, five gates) is consistently among the weakest performers. The paradox arises because gates amplify model-dependent variance: when the LLM can satisfy the gate's criteria, the revise-and-resubmit loop converges quickly; when it cannot, the loop escalates cost without improving quality. We quantify this failure mode in the efficiency analysis (\S\ref{sec:efficiency}).

\subsubsection{Per-Task Heatmap}

\begin{figure}[t]
\centering
\includegraphics[width=\textwidth]{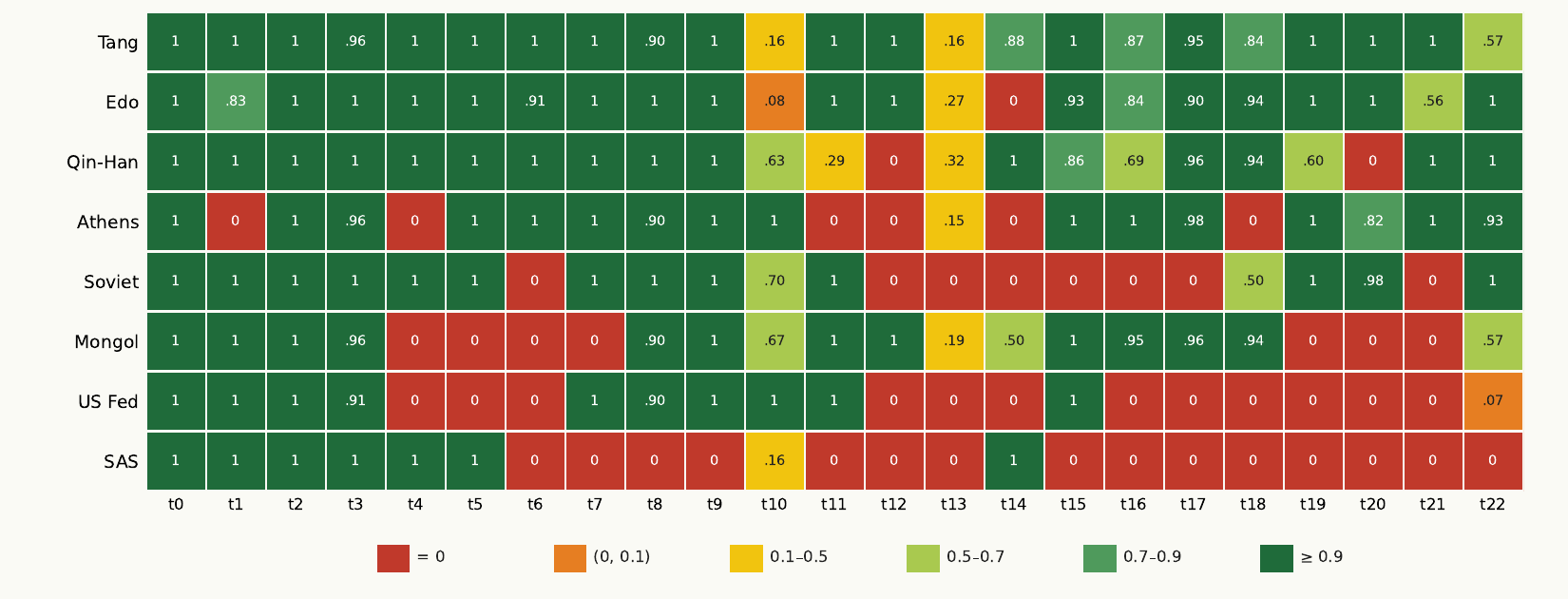}
\caption{Per-task PinchBench score heatmap (MiniMax M2.5). Each cell represents one (institution, task) pair, colored by score ratio. Tang Three Departments achieves the broadest task coverage with 0 zero-score tasks, though it scores low on multi-step workflow (t10) and image generation (t13). The SAS baseline fails on 15 of 23 tasks, confirming that governance scaffolding is essential for complex tasks. Task indices preserve the PinchBench release ordering.}
\label{fig:heatmap}
\end{figure}

Figure~\ref{fig:heatmap} visualizes the per-task score distribution for MiniMax M2.5 across all institutions. The heatmap reveals that institutional advantages are task-dependent: Tang achieves the broadest task coverage with 0 zero-score tasks and perfect scores on 14 of 23 tasks, while SAS---despite succeeding on simple retrieval tasks---fails on 15 of 23 tasks where multi-agent coordination is essential.

\subsection{Efficiency Analysis}
\label{sec:efficiency}

Table~\ref{tab:efficiency} reports efficiency metrics on PinchBench for all three models. We focus on the relationship between structural complexity and cost.

\begin{table}[t]
\caption{Efficiency metrics on PinchBench. \emph{Score}: mean success rate (\%). \emph{Steps}: average runtime steps per task. \emph{Tokens}: average tokens per task ($\times 10^3$). \emph{Zeros}: number of tasks with score $= 0$ (out of 23).}
\label{tab:efficiency}
\centering
\begin{adjustbox}{max width=\textwidth}
\begin{tabular}{@{}l cc cc cc cc cc cc@{}}
\toprule
& \multicolumn{4}{c}{MiniMax M2.5} & \multicolumn{4}{c}{Kimi K2.5} & \multicolumn{4}{c}{Gemini 2.5 Flash} \\
\cmidrule(lr){2-5} \cmidrule(lr){6-9} \cmidrule(lr){10-13}
Institution & Score & Steps & Tok. & Zeros & Score & Steps & Tok. & Zeros & Score & Steps & Tok. & Zeros \\
\midrule
SAS  & 31.1 & 1.0 & 9 & 15 & 21.7 & 1.0 & 4 & 18 & 63.8 & 1.0 & 241 & 4 \\
Qin-Han        & 79.5 & 4.0 & 81 & 2 & 30.2 & 3.8 & 15 & 16 & 65.3 & 4.0 & 253 & 4 \\
Soviet         & 61.6 & 4.6 & 90 & 8 & 8.7 & 4.8 & 16 & 21 & 63.4 & 4.9 & 407 & 4 \\
Mongol         & 59.3 & 5.3 & 93 & 7 & 67.3 & 3.5 & 17 & 5 & 46.3 & 5.1 & 305 & 9 \\
Athens         & 68.4 & 3.0 & 45 & 6 & 12.9 & 2.0 & 8 & 20 & 57.9 & 2.7 & 151 & 8 \\
Edo Bakuhan    & 83.7 & 3.0 & 48 & 1 & 12.3 & 3.0 & 6 & 20 & 87.7 & 3.0 & 216 & 0 \\
\midrule
Tang           & 88.2 & 5.9 & 92 & 0 & 21.6 & 26.7 & 89 & 18 & 27.2 & 5.6 & 235 & 15 \\
US Federal     & 43.0 & 6.8 & 123 & 12 & 13.0 & 6.7 & 22 & 20 & 51.1 & 5.2 & 410 & 9 \\
\bottomrule
\end{tabular}
\end{adjustbox}
\end{table}

\paragraph{The Pareto frontier.}
Edo Bakuhan emerges as the most efficient institution: on MiniMax, it achieves 83.7\% with only 3.0 steps and 48K tokens per task, with just 1 zero-score task. On Gemini, it reaches the overall best 87.7\% with 3.0 steps, 216K tokens per task, and 0 zero-score tasks. Its autonomous cluster pattern achieves high parallelism with minimal coordination overhead.

\paragraph{Gate density and the gate-loop failure mode.}
The efficiency gap between gated and non-gated institutions in Table~\ref{tab:efficiency} is striking: Tang and US Federal consume the most steps and tokens yet do not consistently lead in score. To diagnose this, we examine gate density $\rho$ as a structural predictor. On Kimi, Tang ($\rho = 0.17$) enters catastrophic \emph{gate-loop explosion}: the Menxia gate rejects proposals up to 15 consecutive times per task, producing an average of 26.7 steps and 89K tokens per task---while achieving only 21.6\% overall score. Trace analysis shows that 18 of 23 tasks hit the step budget ceiling (32 steps) without the gate ever approving. On MiniMax, the same topology performs well (5.9 steps, 88.2\% score), demonstrating that the gate-loop risk is a function of the LLM backend's ability to satisfy the gate's approval criteria. The US Federal system ($\rho = 0.56$) avoids loop explosion because its gates use \texttt{reject $\to$ terminal} semantics (outright rejection rather than revision), but this binary accept/reject pathway yields consistently low scores across all models (43.0\%, 13.0\%, 51.1\%).

These results identify a \emph{gate loop} failure mode: when an LLM backend has difficulty satisfying a gate's approval criteria, the reject-revise cycle amplifies step count without improving output quality. Gate density $\rho$ serves as a structural predictor of this risk, and its severity depends critically on the interaction between $\rho$ and the model's instruction-following capability.

\paragraph{Zero-score analysis.}
The distribution of complete failures (zero-score tasks) reveals qualitatively different failure modes. SAS's 15 zeros (MiniMax) reflect raw capability limitations---a single agent without governance scaffolding cannot handle complex tasks. In contrast, US Federal's 12 zeros arise from gate-induced termination: proposals rejected by any of the five sequential gates are terminated rather than revised. Tang's 0 zeros on MiniMax confirm that its error-correction loop achieves near-universal task completion when the model can satisfy the gate, while Tang's 18 zeros on Kimi demonstrate that the same mechanism becomes pathological when the model cannot.

\subsection{Validation on ClaweBench}
\label{sec:clawe-results}

To test the generalizability of our findings, we evaluate all (model, institution) pairs on ClaweBench, a more challenging multi-step benchmark.

\begin{table}[t]
\caption{ClaweBench task success rate (\%) across three models and eight governance specifications.}
\label{tab:clawe}
\centering
\begin{adjustbox}{max width=\textwidth}
\begin{tabular}{@{}llccc@{}}
\toprule
& Institution & MiniMax M2.5 & Kimi K2.5 & Gemini 2.5 Flash \\
\midrule
\multirow{3}{*}{\rotatebox[origin=c]{90}{\scriptsize Pipeline}}
& Single Agent System (baseline) & 45.0 & 47.3 & 20.0 \\
& Qin-Han Commandery-County & \textbf{49.8} & 55.0 & \textbf{25.4} \\
& Soviet Party-State & 47.1 & \textbf{55.5} & 23.6 \\
& Mongol Empire & 46.7 & 51.4 & 23.6 \\
\midrule
\multirow{2}{*}{\rotatebox[origin=c]{90}{\scriptsize Gated}}
& Tang Three Departments & 44.9 & 49.5 & 21.9 \\
& US Federal & \underline{37.9} & 51.5 & \underline{15.7} \\
\midrule
& Edo Bakuhan (cluster) & 42.6 & \underline{47.8} & 16.7 \\
\midrule
& Athenian Democracy (consensus) & 42.5 & 49.8 & 15.9 \\
\midrule
\multicolumn{2}{@{}l}{\emph{Cross-institution mean}} & 44.6 & 51.0 & 20.4 \\
\bottomrule
\end{tabular}
\end{adjustbox}
\end{table}

Table~\ref{tab:clawe} reveals three key differences from PinchBench.

\paragraph{Compressed performance range.}
The gap between the best and worst institution shrinks dramatically: 11.8 pp for MiniMax (Qin-Han 49.8\% vs.\ US Federal 37.9\%), 8.2 pp for Kimi (Soviet 55.5\% vs.\ Bare 47.3\%), and 9.7 pp for Gemini (Qin-Han 25.4\% vs.\ US Federal 15.7\%). This represents a roughly 80\% reduction from the 57--61 pp gaps on PinchBench, suggesting that on sufficiently complex multi-step tasks, the difficulty of the task itself dominates the effect of governance topology.

\paragraph{Ranking reversal.}
The model ranking inverts: Kimi K2.5, which was the weakest model on PinchBench (23.5\% mean), becomes the strongest on ClaweBench (51.0\% mean). MiniMax drops from first to second (44.6\%), and Gemini drops from second to third (20.4\%). This reversal indicates that PinchBench and ClaweBench test fundamentally different capabilities---single-turn tool-use fluency versus multi-step reasoning and planning.

\paragraph{Institutional ranking shifts.}
Edo Bakuhan and Tang---the top two PinchBench performers on MiniMax (83.7\% and 88.2\%)---drop to below average on ClaweBench (42.6\% and 44.9\%). Pipeline institutions (Qin-Han, Soviet) perform relatively better. This suggests that the overhead of complex topologies (parallel clusters, gated loops) provides diminishing returns when tasks require extended reasoning chains, where the base model's planning capability matters more than the governance structure. In the micro-society framing, different task regimes demand different institutional wisdom: simple tasks reward error-correcting mechanisms, while complex tasks reward agent autonomy.

\subsection{Discussion}
\label{sec:discussion}

Returning to the three questions posed at the outset: (1)~governance specifications are a first-order performance determinant, with best-vs-worst gaps of 57--61 pp on PinchBench; (2)~institution rankings do \emph{not} generalize---the top-ranked institution differs across every model and benchmark combination; (3)~structural complexity yields diminishing and sometimes negative returns, as gated topologies can amplify cost without improving quality.

\paragraph{Governance principles and design trade-offs.}
Historical institutions whose governance principles address operational bottlenecks---error correction through centralized review, throughput through decentralized execution---show clear benefits in MAS when paired with a capable LLM backend, but the same mechanisms can become pathological when the model cannot satisfy the topology's demands. Gate density $\rho$ captures this trade-off quantitatively: we conjecture that an optimal $\rho$ exists for each (model, task distribution) pair, and that this optimum decreases as task complexity increases. More broadly, the absence of a universally optimal topology confirms that the micro-society perspective is valuable not as a source of a single best design, but as a \emph{structured design space} for governance architectures that would be difficult to discover through ad-hoc engineering.

\paragraph{Toward self-evolving MAS architectures.}
The cross-model ranking instability and task-dependent nature of institutional advantages suggest that no static governance topology can serve all conditions. A natural extension is a \emph{meta-governance} layer that dynamically selects or adapts the governance specification based on real-time signals (e.g., task complexity estimates, gate approval rates, step-count budgets). The historical analogy is apt: real political institutions evolve over centuries in response to external pressures. Enabling similar adaptability in artificial multi-agent systems---where the ``evolutionary pressure'' comes from task performance feedback---is a promising direction for future work.

\section{Conclusion}
\label{sec:conclusion}

We presented \textsc{SocialSystemArena}, a framework that adopts a micro-society perspective on multi-agent collaboration and formalizes seven historical political institutions as declarative governance specifications evaluated on a unified runtime. Our experiments across three LLM backends and two benchmarks demonstrate that governance topology is a first-order determinant of MAS performance, with best-vs-worst gaps exceeding 57 percentage points within a single model. At the same time, no single institution is universally optimal: rankings shift across models, benchmarks, and task types, revealing that the optimal governance structure depends on the interaction between the task, the model, and the topology. Gate density $\rho$ proves to be a useful structural predictor, identifying conditions under which review mechanisms improve quality versus trigger runaway failure loops. These findings validate the micro-society perspective as a productive design lens---historical institutions provide a structured and diverse space of governance architectures that would be difficult to discover through ad-hoc engineering alone.

\paragraph{Limitations and future work.}
We evaluate three commercial LLM backends; generality to open-weight models remains to be verified. Institutions are modeled as static specifications, losing the adaptive dynamics of real institutional evolution. Key open directions include building a \emph{meta-governance} layer that dynamically reconfigures topology based on runtime signals, extending the institutional search space to modern organizational architectures, and developing principled methods for matching governance structures to task regimes and model capabilities.



\bibliographystyle{plainnat}
\bibliography{references}
\appendix

\appendix

\section{Runtime Implementation Details}
\label{app:runtime}

This appendix provides details on three runtime components deferred from Section~\ref{sec:runtime}: the prompt assembly mechanism, feature plugins, and the LLM adapter protocol.

\subsection{Prompt Assembly}
\label{app:prompt}

The runtime constructs each agent's input through a four-layer prompt assembly:
\begin{enumerate}
    \item \textbf{Soul prompt} ($\sigma_i$): A per-agent markdown file encoding institutional persona, behavioral constraints, and output format requirements.
    \item \textbf{Stage context}: Task state, execution history, and any shared-state variables injected by feature plugins.
    \item \textbf{Tool descriptions}: The set of tools available to the agent at the current stage, serialized as function schemas.
    \item \textbf{Format instructions}: Pattern-specific output constraints (e.g., gate stages must emit an \texttt{approve}/\texttt{reject} decision; consensus voters must emit a \texttt{vote} field).
\end{enumerate}
This layered design separates institutional norms (layer~1) from execution context (layers~2--4), allowing the same soul prompt to be reused across different runtime configurations.

\subsection{Feature Plugins}
\label{app:features}

Features are composable behavioral modifiers that hook into the runtime's \textsc{BeforeStage}/\textsc{AfterStage} lifecycle (Algorithm~\ref{alg:runtime}, lines~10 and~18). Six plugins are provided:

\begin{itemize}
    \item \textbf{Monitor}: A side-channel observer agent that receives the execution trace after each step and can inject warnings or force decisions (used by Qin-Han's \emph{yushi} and Edo's \emph{metsuke}).
    \item \textbf{SharedState}: Maintains a key-value store accessible to all agents, enabling cross-stage information propagation without modifying the message-flow topology.
    \item \textbf{SystemProtocol}: Injects system-level constraints (e.g., budget limits, format requirements) into every agent's context.
    \item \textbf{EmergencyHandler}: Detects error conditions (e.g., tool failures, timeout) and triggers fallback transitions.
    \item \textbf{LoopGuard}: Tracks gate rejection counts and forces an \texttt{approve} decision after a configurable number of consecutive rejections, preventing infinite gate loops.
    \item \textbf{HumanConfirmation}: Optionally pauses execution at designated stages and solicits human input before proceeding.
\end{itemize}

Features are declared in the governance specification's \texttt{features} field and instantiated at compile time. Multiple features compose via ordered execution of their hooks.

\subsection{LLM Adapter Protocol}
\label{app:adapter}

The runtime communicates with LLM backends through a unified adapter interface that abstracts provider-specific APIs:
\begin{itemize}
    \item \texttt{chat(messages, tools, config)} $\to$ \texttt{Response}: Sends a conversation with optional tool schemas and returns the model's response including any tool calls.
    \item \texttt{parse\_decision(response, stage)} $\to$ \texttt{Decision}: Extracts the routing decision from the model's output according to the stage's expected format.
\end{itemize}
Each backend (MiniMax M2.5, Kimi K2.5, Gemini 2.5 Flash) implements this interface with provider-specific serialization. The adapter handles retries, rate limiting, and token counting uniformly, ensuring that differences in task performance reflect only the LLM's capabilities and the governance topology---not adapter-level implementation variance.

\section{Athenian Democracy Case Study}
\label{app:athens}

The Athenian Democracy instantiates the consensus pattern, modeling the direct-democratic institutions of 5th-century BCE Athens.

\begin{figure}[h]
\centering
\includegraphics[width=\textwidth]{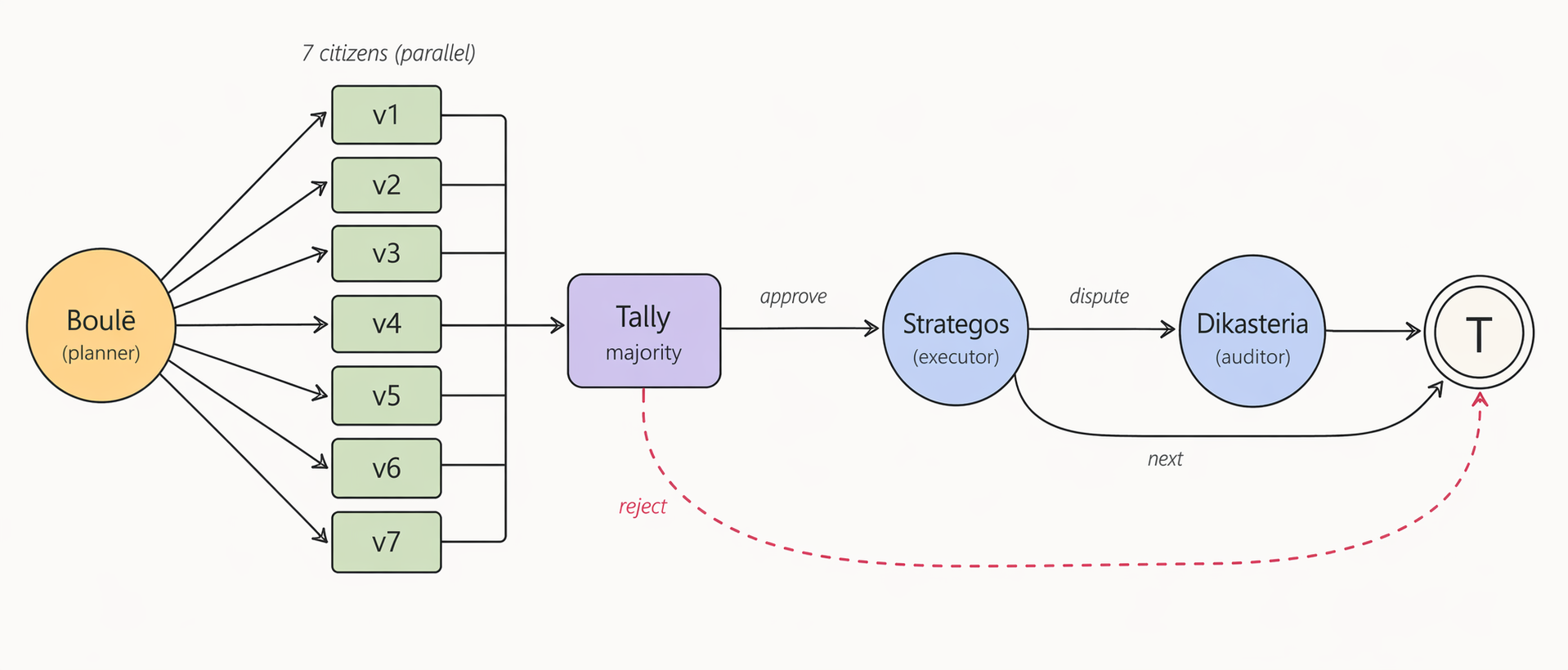}
\caption{Athenian Democracy topology. Seven citizen-voters with distinct ideological biases vote in parallel; the Dikasteria auditor activates only when the executor raises a dispute.}
\label{fig:athens}
\end{figure}

Seven citizen-voters with distinct ideological biases (e.g., fiscal conservative, security realist, civil libertarian) vote in parallel with majority aggregation ($\theta = 0.5$, \texttt{error\_handling = abstain}). The Dikasteria auditor activates only when the Strategos executor raises a \texttt{dispute} decision, modeling retrospective judicial review rather than routine oversight.

\newpage

\end{document}